\documentclass[letterpaper, 10 pt, journal]{IEEEtran}  
\usepackage{amsmath}
\usepackage{graphicx}
\usepackage{amsfonts}
\usepackage{amssymb}
\usepackage{subcaption}
\usepackage{hyperref}
\usepackage{tikz}
\usepackage{booktabs}
\usepackage{cite}
\usepackage{xcolor}
\hypersetup{
    colorlinks=false,
    urlcolor=black,
    pdfborder={0 0 0}
}
\tikzset{imglabel/.style={
    fill=white, 
    fill opacity=0.7, 
    text opacity=1, 
    anchor=north west, 
    inner sep=1.5pt, 
    font=\small\bfseries}
}
\IEEEoverridecommandlockouts                              

\usepackage{cite}
\title{\LARGE \bf
Visual Prompt Guided Unified Pushing Policy
}

\author{Hieu Bui$^{\ast1}$, Ziyan Gao$^{\ast\dagger2}$, Yuya Hosoda$^{3}$, Joo-Ho Lee$^{\dagger3}$
\thanks{$^{1}$ Graduate School of Information Science and Engineering, Ritsumeikan University, Japan.
        {\tt\small hieubui.sami@gmail.com}
}%
\thanks{$^{2}$ Japan Advanced Institute of Science and Technology (JAIST).\\
        {\tt\small ziyan-g@jaist.ac.jp}
}%
\thanks{$^{3}$ College of Information Science and Engineering, Ritsumeikan University, Japan.
        {\tt\small leejooho@is.ritsumei.ac.jp}
}%
\thanks{$^{\ast}$ Equal contribution
}%
\thanks{$^{\dagger}$ Corresponding authors
}
}

\begin{document}

\maketitle
\thispagestyle{empty}
\pagestyle{empty}

\begin{abstract}
As one of the simplest non-prehensile manipulation skills, pushing has been widely studied as an effective means to rearrange objects.
Existing approaches, however, typically rely on multi-step push plans composed of pre-defined pushing primitives with limited application scopes, which restrict their efficiency and versatility across different scenarios.
In this work, we propose a unified pushing policy that incorporates a lightweight prompting mechanism into a flow matching policy to guide the generation of reactive, multimodal pushing actions.
The visual prompt can be specified by a high-level planner, enabling the reuse of the pushing policy across a wide range of planning problems.
Experimental results demonstrate that the proposed unified pushing policy not only outperforms existing baselines but also effectively serves as a low-level primitive within a VLM-guided planning framework to solve table-cleaning tasks efficiently.


\begin{IEEEkeywords}
Visual Prompting, Imitation Learning, Non-prehensile Manipulation, Task And Motion Planning
\end{IEEEkeywords}
\end{abstract}


\section{INTRODUCTION}
\IEEEPARstart{T}{his} work proposes a unified pushing policy that can produce multimodal pushing behavior in multi-object scenarios to facilitate diverse manipulation tasks. 
On the analytics side, pushing has been studied for decades~\cite{stuber2020let}, and the indeterminacy of the friction and the hybrid nature in dynamics (sticking and sliding modes) typically impose challenges in both modeling \cite{push_mason, push_ellipsoid} and control \cite{push_hogan}. Despite these theoretical difficulties, pushing remains widely used in practice due to its ease of deployment, and it also serves as a testbed for the learning-based model.
On the application side, a variety of heuristic or learning-based policies have been proposed for applying pushing to ease manipulation tasks.
An earlier work by Danielczuk {\it et al.}~\cite{push_aid_CASE2018_linePush_Danielczuk} developed several evaluation metrics to select pushing actions that best facilitate the bin picking process, while Zeng {\it et al.}~\cite{push_aid_IROS2018_grasp_Zeng} models the synergies between the pushing and grasping policies via deep reinforcement learning. In addition, pushing can also be used for object displacement~\cite{push_aid_TASE2022_displace_GAO}, singulation~\cite{push_aid_ISRR2019_singulate_Eitel}, and target retrieval~\cite{push_aid_ICRA2021_targetRetrieve_Huang}.

\begin{figure}
    \centering
    \includegraphics[width=0.96\linewidth]{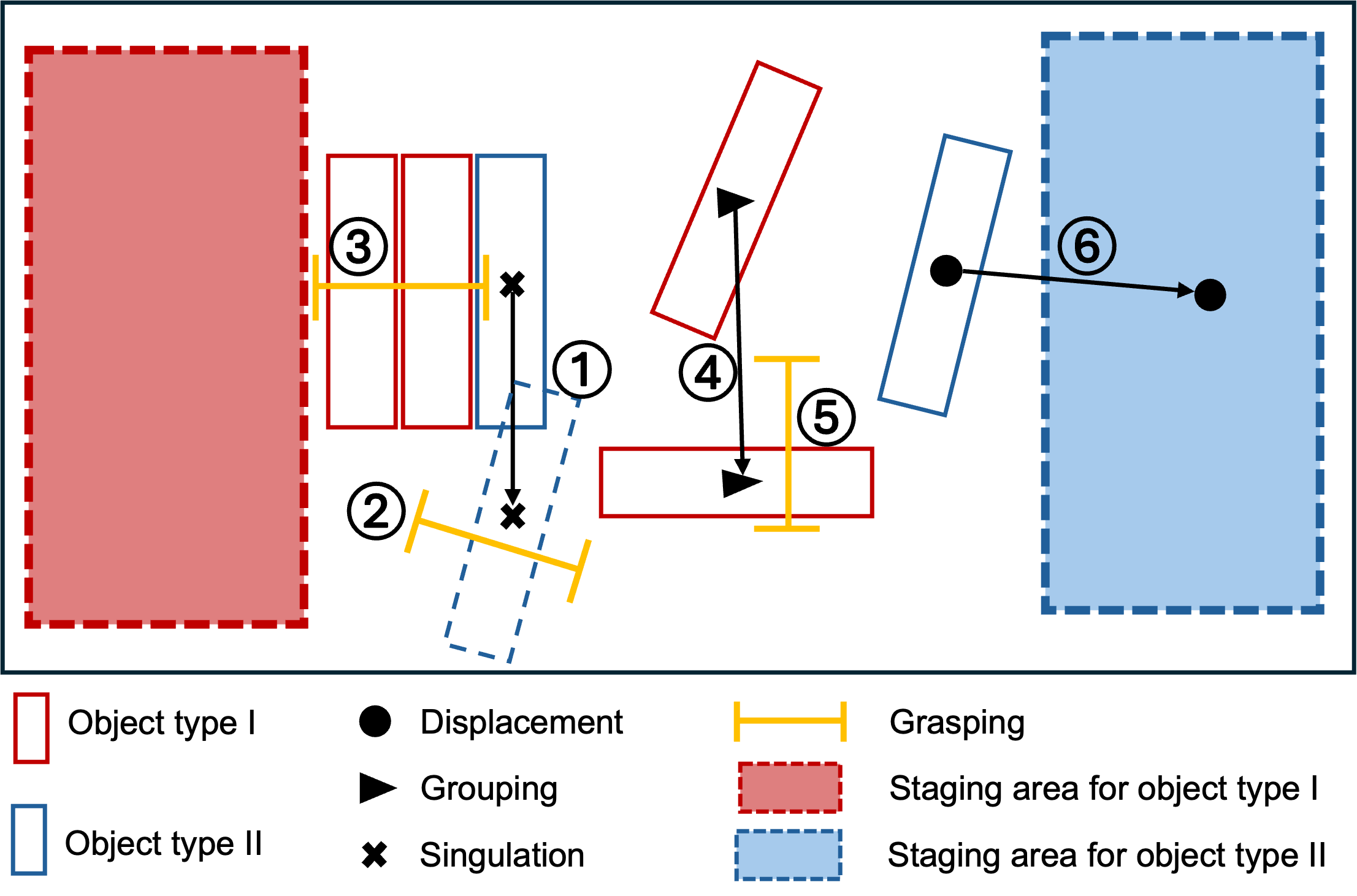}
    \caption{Illustration of a specific table-cleaning task in which all red blocks must be placed in the left staging area, while blue blocks are placed in the right staging area. The numbered annotations indicate one possible sequence of actions considering the feasibility and efficiency.}
    \label{fig:fancy_cleanning}
    \vspace{-10pt}
\end{figure}

Although significant progress has been made in pushing applications, most existing pushing policies remain task-oriented, making them difficult to reuse when the environment or task requirements change. For example, a pushing policy tailored for aiding bin picking may not be effective in a target-retrieval scenario—a task that requires exposing and extracting a specific object from clutter rather than clearing all objects.
Moreover, even within a single task domain, a unimodal pushing policy may not be sufficient to efficiently solve the problem. Yonemaru {\it et al.}~\cite{push_Arxiv2025_Yonemaru} proposed an imitation-learning model that learns only a grouping behavior to facilitate efficient multi-object grasping. However, when objects are already heavily cluttered, grouping becomes suboptimal; instead, a singulation behavior is preferred to declutter the scene and expose graspable items. Therefore, versatile pushing not only enables flexible handling of diverse tasks but also improves efficiency by selectively combining multiple pushing skills according to task requirements and the current scene configuration. 
Fig.~\ref{fig:fancy_cleanning} illustrates an example that highlights the benefits of incorporating multimodal pushing skills to efficiently solve a table-cleaning task. Initially, singulation is applied to isolate a blue block, which is then pick-and-placed via grasping. The third action pick-and-places two red blocks simultaneously. The fourth action groups two red blocks using pushing, followed by a fifth action that pick-and-places the grouped red blocks together. Finally, the sixth action rearranges the remaining blue block into the target region via displacement. On the other hand, the task would fail without a singulation skill, while the execution would become significantly more time-consuming in the absence of grouping or displacement skills.

To advance from unimodal to multimodal pushing policies, in this work, we present a goal-conditioned flow matching policy that encapsulates multiple pushing skills acquired from human demonstration data. We consider three types of non-prehensile skills as mentioned in Fig.~\ref{fig:fancy_cleanning}: displacement, grouping, and singulation. In the displacement task, the objective is to push the target object to a specified position. In the grouping task, the goal is to bring multiple objects together, facilitating potential multi-object grasping. In the singulation task, the aim is to isolate the target object from surrounding clutter. 

A distinguishing feature of our model is the integration of a prompting mechanism to guide multimodal pushing action generation. Specifically, the prompting mechanism combines keypoints from the camera frame to indicate either the object of interest or the desired target position, and a task specifier encoding which type of pushing skill to apply. This feature not only facilitates the generation of different task-oriented pushing action but also enables the integration with high-level task planners to re-use the learned pushing policies for solving manipulation tasks effectively.

Both the training and testing of the proposed unified pushing policy are conducted on the real platform.  Experimental results demonstrate that, in comparison with single-task baselines and goal-image–based models, the proposed approach achieves superior or at least comparable performance across all evaluated tasks. Furthermore, additional experiments are conducted to validate its generalization capability to previously unseen objects. Finally, we demonstrate the integration of the proposed policy with a Vision-Language Model-based planner to solve sequential manipulation planning tasks, thereby highlighting the policy’s reusability and effectiveness as a low-level manipulation primitive. Our contribution can be summarized as follows:
\begin{itemize}
    \item We introduce a unified pushing policy based on flow matching that integrates three distinct skills—displacement, grouping, and singulation—into a single model.
    \item We propose a prompting mechanism that combines visual keypoints with task specifiers to guide pushing behaviours.
    \item  We validate our method on a real robotic arm, demonstrating that the unified policy outperforms existing baselines, and showing its utility as a low-level primitive in a VLM planning framework.
\end{itemize}



\section{RELATED WORK}
In multi-object scenarios, pushing has been employed to address general object rearrangement problem~\cite{push_ils_huang, push_kinodynamic_planning_ren}. For instance,  Huang~\textit{et al.}~\cite{push_ils_huang} proposed a meta-heuristic search framework that samples a sequence of open-loop straight-line pushing actions by simulating the robot-object and object-object interactions. To eliminate the uncertainties accompanied with the robot-object interactions, Ren~\textit{et al.}~\cite{push_kinodynamic_planning_ren} proposed a unified planning framework that interleaves the planning and robot execution. However, these works assume the accessible to the full state of the known objects, limiting their applications for unknown objects in real settings. Alternatively, learning-based methods have been proposed to directly sample task-oriented pushing actions without explicit search or planning. These methods typically focus on specific sub-tasks such as target retrieval~\cite{push_aid_ICRA2021_targetRetrieve_Huang, push_invisible_yang}, object singulation~\cite{push_aid_ISRR2019_singulate_Eitel, push_singulate_dong, push_singulate_berscheid}, and bin picking~\cite{push_aid_CASE2018_linePush_Danielczuk, push_aid_IROS2018_grasp_Zeng, push_language_guide_Zhao}. However, they often do not explicitly specify the target object pose and may lack generalizability across task types.

For pushing interaction, straight-line pushing with finite displacement is widely adopted across various tasks~\cite{push_aid_ISRR2019_singulate_Eitel,push_singulate_dong,push_aid_ICRA2021_targetRetrieve_Huang,push_aid_CASE2018_linePush_Danielczuk}. Beyond straight-line motions, curvilinear pushing action have also been explored~\cite{push_IROS2021_Sakamoto}, particularly for grouping nearby objects. However, the reliance on predefined action primitives often limits task efficiency due to their lack of flexibility, open-loop execution, and sensitivity to object motion uncertainties. To overcome these limitations, Wang~\cite{push_uno_wang} introduced a model predictive control framework combined with a smoothing algorithm to enable reactive pushing. 
More recently, many works have begun adopting generative models like diffusion \cite{chi2025diffusion} or flow matching \cite{push_rouxel2024flow} to learn pushing behaviors from data instead of using analytical methods. Yonemaru~\cite{push_Arxiv2025_Yonemaru} employed a diffusion policy to learn pushing actions directly from human demonstrations. One close work is done by Wang~\textit{et al.}~\cite{push_TRO2025_HDP_Wang} that incorporates the contact prompts defined on the image frame to explicitly guide the diffusion policy. Different from these works, we develops a flow matching-based pushing policy, that incoporates visual prompts and task specifiers to specify the object of interest or target position.

Our work is based on goal-conditioned imitation learning, a framework for training policies that are steerable towards a specified goal state.
Natural language \cite{goal_mees2022matters} is a common modality for goal specification  but spatially ambiguous, while goal-image  \cite{goal_reuss2023goal} can provide precise spatial details but is often over-specified and difficult to obtain during inference. Other studies explored using multimodal goal with languages and images \cite{goal_reuss2024multimodal}, and alternative, more abstract goal representations like sketches \cite{goal_sundaresan2024rt} or visual prompts. 
Methods such as PIVOT \cite{prompt_nasiriany2024pivot}, MOKA \cite{prompt_fangandliu2024moka}, and RoVI \cite{prompt_li2025robotic} leverage VLMs to interpret visual annotations (arrows, points, $\dots$) on images to generate motion plans. While demonstrating strong generalization, these methods rely on querying large foundation models, making them computationally expensive to deploy.  Closest to our work, Muttaqien \textit{et al.} \cite{prompt_muttaqien2025visual} propose an Action Chunking Transformers (ACT) policy conditioned on RGB image with bounding box overlay for pick-and-place tasks. In contrast, our method adopts a more lightweight representation, using raw pixel coordinates similar to SAM \cite{kirillov2023segment} and a task specifier to guide a flow matching-based policy.

\section{METHOD}
\begin{figure*}
    \centering
    \includegraphics[width=0.95\linewidth]{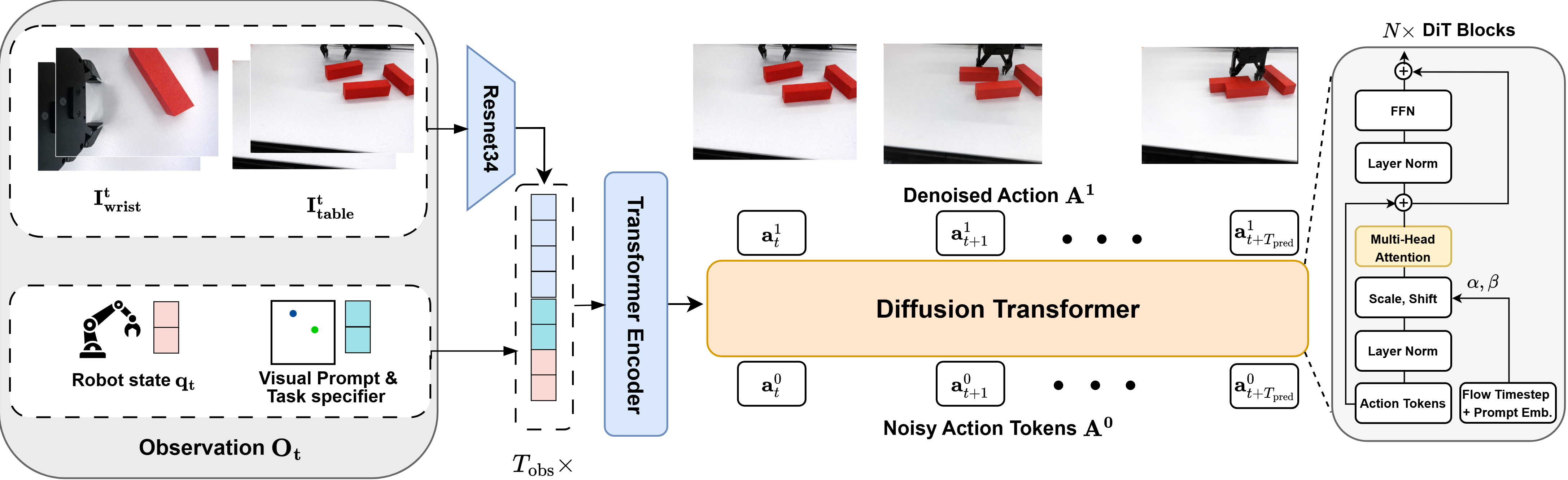}
    \caption{\textbf{Model Architecture.} The input consists of the visual prompt and the latest $T_{\text{obs}}$ steps of image data and robot proprioception. The policy is parameterized by a Diffusion Transformer with alternating self-attention and cross-attention DiT blocks to denoise action tokens $\mathbf{A^0}$ into executable trajectories $\mathbf{A^1}.$
} 
    \label{fig:architecture}
    \vspace{-10pt}
\end{figure*}



\subsection{Problem Formulation}
\label{subsec:problem}


We consider a robotic pushing problem in a partially observable, multi-object environment. A variable number of $N$ rigid objects are stably placed on a planar surface. Each object $i$ has a planar pose $(\mathbf{p}^i_t, \theta^i_t)\in SE(2)$ at time $t$, representing its $2$D position and orientation with respect to the robot's base. The system state is defined as:
\[
s_t = \{ (\mathbf{p}^i_t, \theta^i_t)\}_{i=1}^N, \quad s_t \in \mathcal{S}_N, \quad \text{where} \quad \mathcal{S}_N = SE(2)^N
\] 
At the beginning of each episode, the robot is provided with a task $\psi$ specified by the prompting mechanism in Section~\ref{section:prompting}, which remains fixed throughout the interaction. Following the setting in~\cite{push_Arxiv2025_Yonemaru}, we use a wrist-mounted camera and a table-mounted camera as the visual observation, denoted as $\mathbf{I}_t = [\mathbf{I}^{\text{table}}_t, \ \mathbf{I}^{\text{wrist}}_t]$ as can be seen in Fig.~\ref{fig:architecture}. At each time step $t$, the robot receives an observation $\mathbf{o}_t$, which consists of a visual observation $\mathbf{I}_t$ and the proprioceptive state represented by pose vector $\mathbf{q}_t \in \mathbb{R}^7$, representing the end-effector pose in base frame:
    \begin{equation}
\mathbf{q_t} = \begin{bmatrix} \mathbf{p}_{\text{eef}}^{\text{base}} \ \boldsymbol{\eta}_{\text{eef}}^{\text{base}} \end{bmatrix}^\top = \begin{bmatrix} x \ y \ z \ \eta_w \ \eta_x \ \eta_y \ \eta_z \end{bmatrix}^\top,
    \end{equation}
where $\mathbf{p}_{\text{eef}}^{\text{base}} \in \mathbb{R}^3$ is the Cartesian position and $\boldsymbol{\eta}_{\text{eef}}^{\text{base}} \in \mathbb{R}^4$ is the unit quaternion orientation.
The goal is to generate a reactive pushing trajectory to rearrange object(s) such that the final object configuration satisfies the corresponding goal predicate $\mathcal{G}_\psi$. Formally, the objective can be represented as:
\begin{equation}
\begin{aligned}
\textbf{Given: } & \quad \mathbf{o}_t, \psi \\
\textbf{Find: } & \quad \mathbf{a}_{0:T-1} \text{ such that } s_T = f(s_0, \mathbf{a}_{0:T-1}) \text{ and } \\
& \quad\mathcal{G}_\psi(s_T) = \texttt{True}
\end{aligned}
\end{equation}
Here, $f$ is the environment transition function, $\mathbf{a}_t\in \mathbb{R}^7$ is the action to be executed at time $t$, and $\mathcal{G}_\psi(s_T)$ is a goal predicate that will be introduced in Section~\ref{section:goal predicate}. We use the same representation for $\mathbf{q}_t$ and $\mathbf{a}_t$.

\subsection{Goal Predicates}
\label{section:goal predicate}
To better convey the idea, we define two oracle mappings: \[
\texttt{Obj}(\mathbf{I}_0^{table},\ \mathbf{u}_g) \rightarrow i,\quad
\texttt{Pos}(\mathbf{I}_0^{table},\ \mathbf{u}_g) \rightarrow \mathbf{p}^* \in \mathbb{R}^2
\]
Specifically, $\mathbf{I}_0^{table}$ represents the initial observation from the table-fixed camera. $\mathbf{u}_g$ represents a pixel coordinate \textit{w.r.t} $\mathbf{I}^{table}_0$. $\texttt{Obj}(\mathbf{I}_0^{table}, \mathbf{u}_g)$ returns the index of the object whose image projection contains $\mathbf{u}_g$, while $\texttt{Pos}(\mathbf{I}_0^{table}, \mathbf{u}_g)$ indicates a position on the tabletop \textit{w.r.t} the table fixed camera. Then the task-relevant goal predicates are defined as follows:
\begin{itemize}
    \item \textbf{Displacement:} Move a target object to a desired location $\texttt{POS}(\mathbf{I}_0^{table}, \mathbf{u}_g)$.
    \begin{equation}
\|\mathbf{p}^{\texttt{Obj}(\mathbf{I}_0^{table}, \mathbf{u}_1)}_T - \texttt{Pos}(\mathbf{I}_0^{table}, \mathbf{u}_2)\| < \epsilon
    \end{equation}
    where $\epsilon$ is a distance tolerance.

    \item \textbf{Grouping:} Bring two specified objects close together for potential grasping. Proximity and relative orientation should be satisfied:

    \begin{equation}
    \begin{aligned}
    &\|\mathbf{p}^{\texttt{Obj}(\mathbf{I}_0^{table}, \mathbf{u}_1)}_T - \mathbf{p}^{\texttt{Obj}(\mathbf{I}_0^{table}, \mathbf{u}_2)}_T\| < \epsilon \quad \text{and}  \\
     &|\theta^{\texttt{Obj}(\mathbf{I}_0^{table}, \mathbf{u}_1)}_T - \theta^{\texttt{Obj}(\mathbf{I}_0^{table}, \mathbf{u}_2)}_T| < \delta
    \end{aligned}
    \end{equation}
    \item \textbf{Singulation:} Isolate a target object from clutter. The object must not be in close proximity to any other object:
    \begin{equation}
    \forall j \neq \texttt{Obj}(\mathbf{I}_0^{table}, \mathbf{u}_1),\ \|\mathbf{p}^{\texttt{Obj}(\mathbf{I}_0^{table}, \mathbf{u}_1)}_T - \mathbf{p}^j_T\| > r
    \label{eq:singulation}
    \end{equation}
    where $r$ is a minimum separation distance.
\end{itemize}

The thresholds $\epsilon$, $\delta$, $r$ serve as \emph{nominal descriptors} of the task goals; we emphasize that they are not used as explicit supervision during training or evaluation. Instead, the policy is trained via imitation learning from human demonstrations and learns to satisfy these task objectives implicitly through data. As such, these predicates define the conceptual structure of the task, while the model’s behavior approximates goal satisfaction through learned patterns rather than hard constraints.

\subsection{Prompting mechanism}
\label{section:prompting}
\begin{figure}[!ht]
    \centering
    
    \begin{subfigure}{\linewidth}
        \centering
        \includegraphics[width=0.46\linewidth]{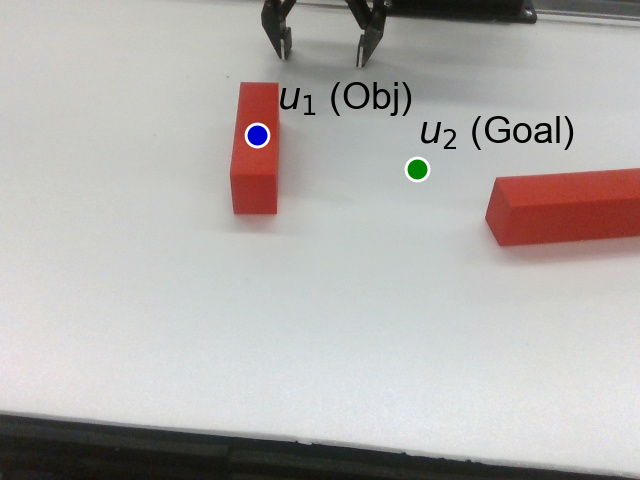}
        \vspace{5pt}
        \includegraphics[width=0.46\linewidth]{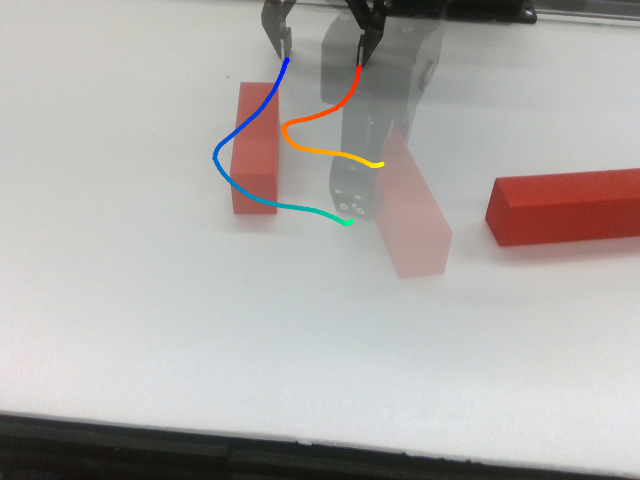}
        \caption{Displacement}

    \end{subfigure}
    
    
    \begin{subfigure}{\linewidth}
        \centering
        \includegraphics[width=0.46\linewidth]{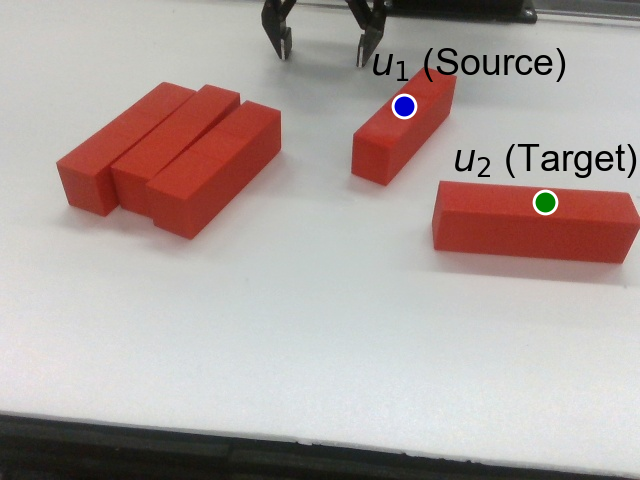}
        \vspace{5pt}
        \includegraphics[width=0.46\linewidth]{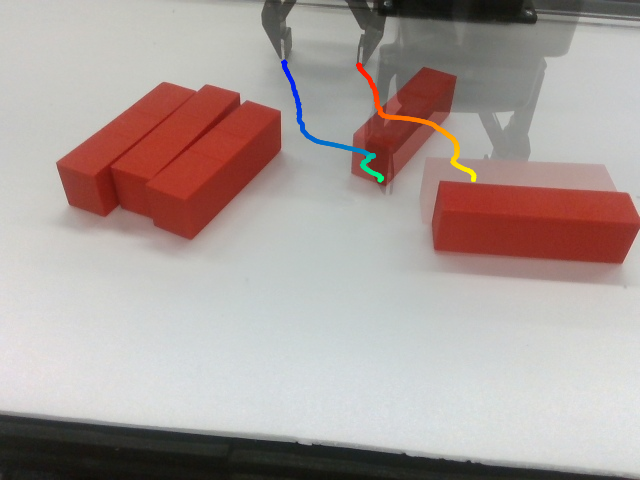}
        \caption{Grouping}

    \end{subfigure}
    
    
    \begin{subfigure}{\linewidth}
        \centering
        \includegraphics[width=0.46\linewidth]{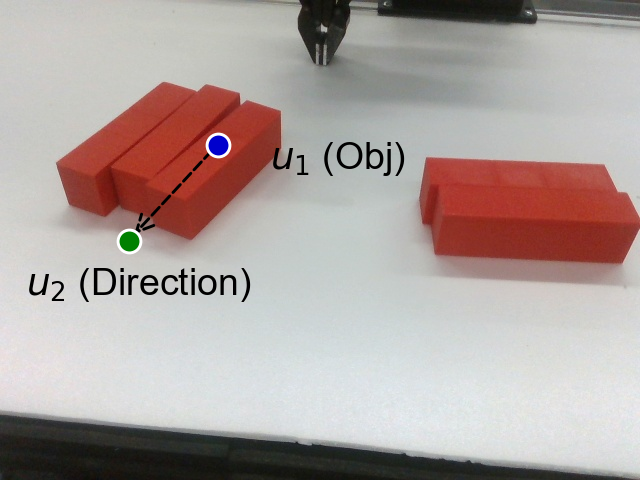}
        \vspace{5pt}
        \includegraphics[width=0.46\linewidth]{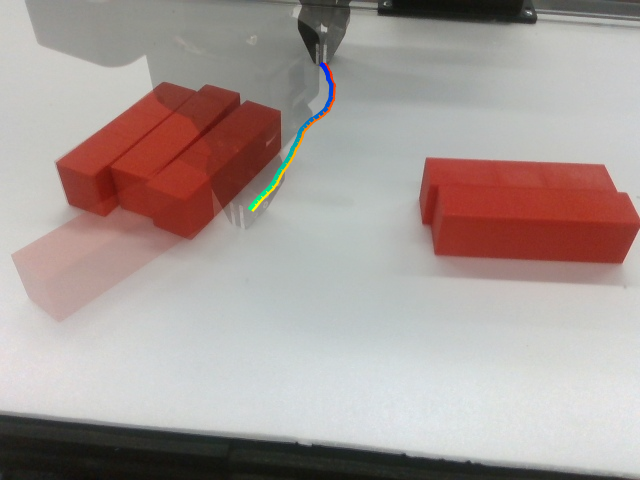}
        \caption{Singulation}

    \end{subfigure}
    
    \caption{Illustration of the visual prompt. \textbf{Left:} Input prompts consisting of two points: $\mathbf{u}_1$ (blue) and $\mathbf{u}_2$ (green). \textbf{Right:} Policy-generated action trajectories. The warm and cold colors represent the trajectory of the right and left finger of the gripper.}
    \label{fig:prompt}
    \vspace{-10pt}
\end{figure}
\paragraph{Prompting Mechanism.}
In this work, we introduce a prompting mechanism as an external interface for specifying a task $\psi$. The prompting mechanism consists of (i) a \emph{task specifier} $\psi_{\text{task}} \in \{\texttt{displacement}, \texttt{grouping}, \texttt{singulation}\}$, and (ii) \emph{visual prompts} defined on $\mathbf{I}^{\text{table}}_0$. Specifically, the visual prompts are represented by a pair of pixel coordinates $[\mathbf{u}_1, \mathbf{u}_2]$ on $\mathbf{I}^{\text{table}}_0$. The semantic interpretation of these visual prompts is conditioned on the task specifier $\psi_{\text{task}}$, as detailed below:
\begin{itemize}
    \item \textbf{Displacement} ($\psi_{\text{task}} = \texttt{displacement}$):
    The pixel $\mathbf{u}_1$ specifies the object to be pushed such that $\texttt{Obj}(\mathbf{I}^{\text{table}}_0, \mathbf{u}_1)$ denotes the source object. The pixel $\mathbf{u}_2$ indicates the desired target location, represented by $\texttt{Pos}(\mathbf{I}^{\text{table}}_0, \mathbf{u}_2)$, to which the source object should be displaced.

    \item \textbf{Grouping} ($\psi_{\text{task}} = \texttt{grouping}$):
    The pixel $\mathbf{u}_1$ identifies the source object, i.e., $\texttt{Obj}(\mathbf{I}^{\text{table}}_0, \mathbf{u}_1)$, while $\mathbf{u}_2$ specifies the target object via $\texttt{Obj}(\mathbf{I}^{\text{table}}_0, \mathbf{u}_2)$, indicating that the source object should be pushed to form a group with the target object.

    \item \textbf{Singulation} ($\psi_{\text{task}} = \texttt{singulation}$):
    The pixel $\mathbf{u}_1$ denotes the source object, i.e., $\texttt{Obj}(\mathbf{I}^{\text{table}}_0, \mathbf{u}_1)$. The vector $\texttt{Pos}(\mathbf{I}^{\text{table}}_0, \mathbf{u}_2) - \texttt{Pos}(\mathbf{I}^{\text{table}}_0, \mathbf{u}_1)$ defines the pushing direction used to displace the source object so as to satisfy the singulation condition in Eq.~\ref{eq:singulation}.
\end{itemize}
The sub-figures in the left column of Fig.~\ref{fig:prompt} illustrate representative visual prompts for each task type. Given the visual prompts and the corresponding task specifier, the pushing trajectory is generated as shown in the right column of Fig.~\ref{fig:prompt}.

During both training and inference, the task specification $\psi$ is encoded as a goal vector $\mathbf{g}$ derived from the task specifier and visual prompts. Specifically, the task specifier is first encoded as a one-hot vector, while the visual prompts $[\mathbf{u}_1, \mathbf{u}_2]$ are normalized to the range $[0,1]$. For the \texttt{displacement} and \texttt{grouping} tasks, the goal vector $\mathbf{g}$ is constructed by concatenating the one-hot task encoding with the normalized visual prompts. In contrast, for the \texttt{singulation} task, the goal vector $\mathbf{g}$ is formed by concatenating the one-hot task encoding, the first normalized visual prompt, and the normalized pushing direction $\frac{\mathbf{u}_2 - \mathbf{u}_1}{\lVert \mathbf{u}_2 - \mathbf{u}_1 \rVert}$,
which explicitly encodes the desired displacement direction of the source object.

\subsection{Imitation Learning Model}
\subsubsection{Flow Matching Policy}
 Flow Matching \cite{lipman2022flow} is a class of generative models based on optimal transport theory that aims to estimate a time-dependent vector field $v^\tau$ that continuously transforms samples from a \textit{source} distribution $p_0$ to samples of a \textit{destination} distribution $p_1$. We adopt Flow Matching over Diffusion \cite{chi2025diffusion} because it constructs straight-line optimal transport paths from noise to action. This simplifies the training objective and enables faster inference for high-frequency control. Given a dataset of $N$ demonstrations $\mathcal{D} = \{\mathbf{g}^n, \{\mathbf{o}_t^n, \mathbf{a}_t^n\}_{t=1}^{T(n)}\}_{n=1}^N$, each of length $T(n)$, our goal is to learn a goal-conditioned policy $\pi(\mathbf{A_t} | \mathbf{O_t}, \mathbf{g})$ by employing flow matching to model the conditional distribution $p(\mathbf{A_t}|\mathbf{O_t}, \mathbf{g})$, where $\mathbf{A_t} = [\mathbf{a}_t, \mathbf{a}_{t+1}, \dots, \mathbf{a}_{t+T_{\text{pred}}-1}]$ is a chunk of $T_{\text{pred}}$ future, and $\mathbf{O_t} = [\mathbf{o}_{t-T_{\text{obs}}+1}, \mathbf{o}_{t-T_{\text{obs}}+2}, \dots, \mathbf{o}_t]$ is a history of $T_\text{obs}$ observations. During training, given a ground-truth action chunk $\mathbf{A}_t^1$, a flow matching step $\tau\in[0, 1]$, and Gaussian noise $\mathbf{A}_t^0 \sim \mathcal{N}(\mathbf{0}, \mathbf{I})$, we can calculate a noisy action chunk  $\mathbf{A}^\tau_t = \tau\mathbf{A}_t^1 + (1-\tau)\mathbf{A}_t^0$. The vector field $v^\tau$ can be learned by a neural network $v_\theta^\tau(\mathbf{A}_t^\tau, \mathbf{O}_t, \mathbf{g})$ using the  Conditional Flow Matching loss: 
 \begin{equation}
    \begin{split}
        \mathcal{L}_\text{CFM}(\theta) &= \mathbb{E}_{\tau,p_0,p_1}||d\mathbf{A}_t^\tau / d\tau - v_\theta^\tau(\mathbf{A}_t^\tau, \mathbf{O}_t, \mathbf{g})||^2 \\
        &= \mathbb{E}_{\tau,p_0,p_1}||(\mathbf{A}^1_t - \mathbf{A}^0_t) - v_\theta^\tau(\mathbf{A}_t^\tau, \mathbf{O}_t, \mathbf{g})||^2
    \end{split}
    \label{eq:cfm_loss}
\end{equation}
During inference, we generate an action chunk by first sampling $\mathbf{A}^0_t \sim \mathcal{N}(\mathbf{0}, \mathbf{I})$ and use forward Euler integration with $K=5$ steps to integrate the learned vector field:
\begin{equation}
    \mathbf{A}_t^{\tau+\Delta\tau} = \mathbf{A}_t^{\tau} + \Delta\tau v_\theta^\tau(\mathbf{A}_t^\tau, \mathbf{O}_t, \mathbf{g})
    \label{eq:euler}
\end{equation}

\subsubsection{Prompt Guidance}
To enhance the policy's adherence to the prompt, we employ an additional conditioning method used for diffusion and flow models, Classifier-Free Guidance (CFG) \cite{ho2022classifier}. In addition to the goal-conditioned policy $\pi(\mathbf{A_t} | \mathbf{O_t}, \mathbf{g})$, we also implicitly train an unconditioned policy $\pi(\mathbf{A_t} | \mathbf{O_t}, \varnothing)$ by randomly replacing the goal vector $\mathbf{g}$ with a null goal vector $\varnothing$ with a rate of $0.1$. The vector $\varnothing$ is obtained by setting all coordinates in the visual prompt  $\mathbf{u}=\mathbf{0}$, and the task identifier $\psi_{task}$ to a dedicated "null" task category. 

During inference, we modify the predicted vector field $v^\theta_\tau(\mathbf{A}_t^\tau, \mathbf{O}_t, \mathbf{g})$ at each ODE solver step as a linear combination of the conditional and unconditional estimates:
\begin{equation*}
    v_\theta^\tau(\mathbf{A}_t^\tau, \mathbf{O}_t, \mathbf{g}) = \gamma
    v_\theta^\tau(\mathbf{A}_t^\tau, \mathbf{O}_t, \mathbf{g}) + (1-\gamma)v_\theta^\tau(\mathbf{A}_t^\tau, \mathbf{O}_t, \varnothing),
\end{equation*}
where $\gamma$ is the guidance scale. By setting $\gamma > 1$, we push the generated action trajectory toward regions of the distribution that are more strongly aligned with the specified prompt.

\subsubsection{Network Architecture}
Images from both cameras are processed by a shared pretrained Resnet34 backbone with spatial softmax pooling to extract feature maps. These visual features are then flattened and concatenated with the proprioception vector $\mathbf{q}_t$ for each observation timestep, as well as the goal vector $\mathbf{g}$. To capture temporal dependencies over $T_{\text{obs}}$ observation timesteps, we use a standard Transformer Encoder comprised of several self-attention blocks to process the observations, resulting in a sequence of latent observation embeddings. 

To model the vector field $v_\theta$, we use a variant of the DiT \cite{peebles2023dit} architecture, which operates on the noisy action sequence $\mathbf{A}_t^\tau$. The network consists of $N$ alternating self-attention and cross-attention blocks. The input action chunk is projected to the network's hidden dimension with an MLP before entering the network, and after the final DiT block, we apply another MLP to project the activations to the action's dimension to form the predicted vector field. We use a variant of DiT that incorporates conditioning information through adaptive layer normalization (AdaLN) and cross-attention.
The latent observation embeddings are integrated into the network using the cross-attention blocks, allowing the policy to attend to historical visual features when denoising actions. Additionally, we also integrate the current flow timestep $\tau$ and the goal vector $\mathbf{g}$ via  AdaLN. We project the goal vector $\mathbf{g}$ to the DiT network's hidden dimension using a small MLP and add the resulting vector to the sinusoidal positional embedding of $\tau$. This combined embedding is used to regress the scale and shift parameters in the AdaLN block that modulate the activations in the network.

\subsubsection{Implementation Details}
Our implementation is built upon the LeRobot library. Following common literature \cite{chi2025diffusion}, the observation horizon is set to $T_{\text{obs}} = 2$, and the action prediction horizon is  $T_{\text{pred}} = 16$, with an execution horizon of $T_{\text{exec}} = 8$ before replanning. We train the policy for $300$k steps with a batch size of $32$, a learning rate of $1$e-$4$ with $1000$ steps warmup, and a weight decay of $1$e-$6$ using the t
AdamW optimizer. The Resnet34 backbone uses a learning rate one-tenth that of the policy. The DiT network consists of $N=8$ alternating self-attention and cross-attention blocks with a hidden dimension of 512 and 8 attention heads. For the classifier-free guidance, we use a guidance scale of $\gamma = 5.0$ during inference.

\section{Experimental Setup}

\subsection{Hardware Setup}
\begin{figure}[t]
    \centering
    \includegraphics[width=0.9\linewidth]{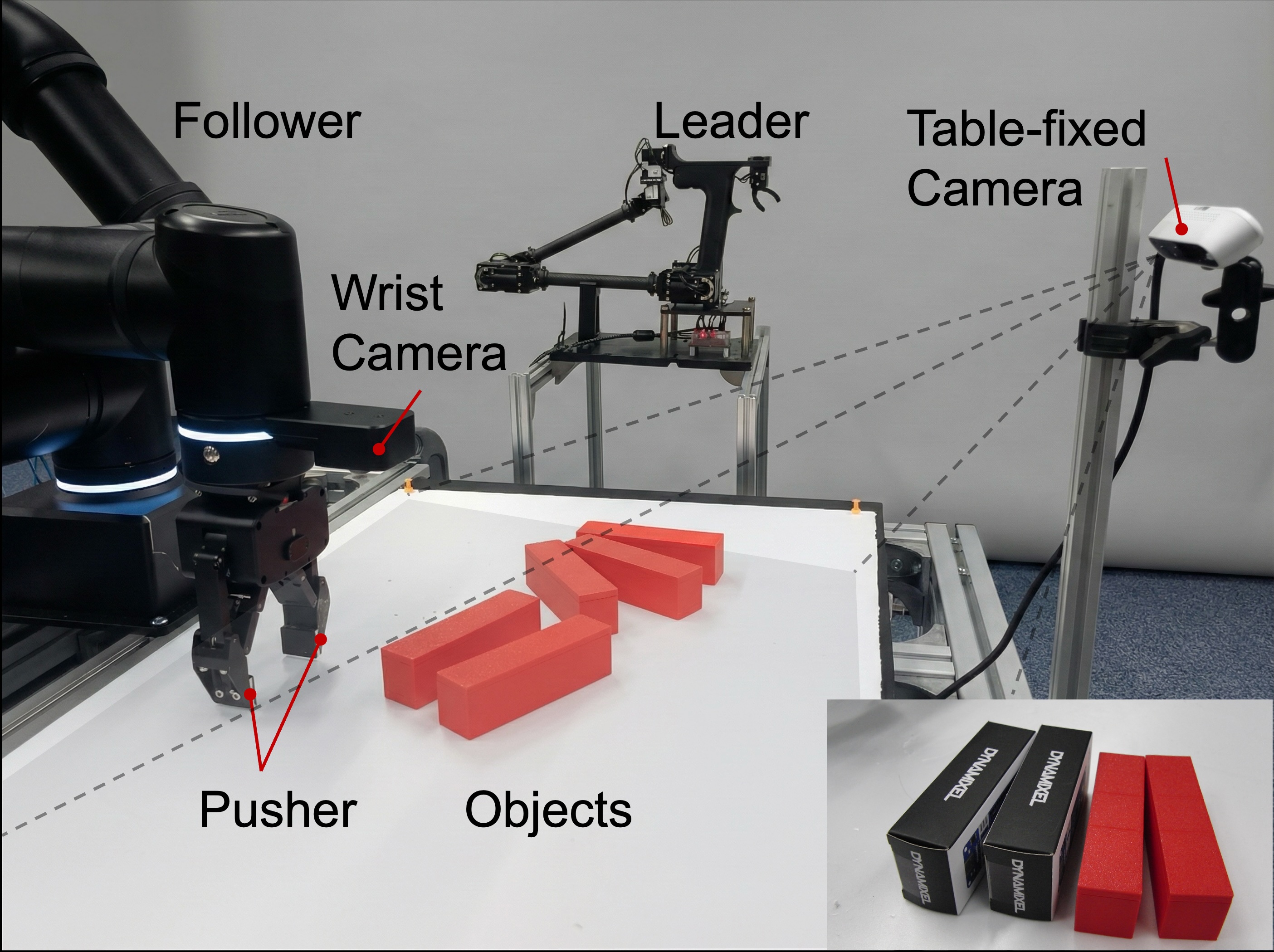}
    \caption{The experimental setup consists of a leader-follower system. The follower (left) is equipped with a wrist-mounted camera and a parallel jaw gripper. The leader device (middle) is used to teleoperate the follower. In the data collection phase, only the red blocks were used, while the embedded figure at the bottom right shows objects used for evaluation.}
    \label{fig:experimental_setting}
    \vspace{-11pt}
\end{figure}

We conduct our experiments on the ROBOTIS OpenManipulator-Y (OM-Y) platform\footnote{OpenManipulator-Y: \url{https://ai.robotis.com/omy/introduction_omy.html}}, as shown in Fig.\ref{fig:experimental_setting}. The system consists of a 6-DOF follower arm and a corresponding leader device for human teleoperation.  During data collection, a human operator performs a task by manipulating the leader device, which the follower reproduces in real-time. 

To maintain consistent pushing dynamics and to reduce the cognitive load on the human operator during teleoperation, we impose kinematics constraints on the follower arm. Following the configuration in \cite{push_Arxiv2025_Yonemaru}, the end-effector's orientation is constrained such that its local $z$-axis\footnote{End-effector's axis:\ \url{https://ai.robotis.com/omy/hardware_omy.html}} remains parallel to the surface. Furthermore,  we fix the height of the end-effector relative to the table, limiting its movements to planar motion.


To capture image data, we use an Intel RealSense D405 camera mounted on the follower's wrist and a table-mounted Intel RealSense D435. Both cameras record $640 \times 480$ video at $15$ FPS, which are resized to $320 \times 240$ for model input. The objects used for experiments are 3D-printed rectangular cuboids measuring $120 \times 30 \times 30$~mm. Data recording is conducted at $10$~Hz. During policy execution, the policy generates action chunks as end-effector's poses; these poses are then converted to joint commands using inverse kinematics and sent to the robot at a fixed rate of $10$~Hz. 

\subsection{Data Collection}
To train the policy and the baselines, we collect a dataset of $550$ expert demonstrations using the leader-follower teleoperation setup described above. A simple graphical user interface is developed to allow the human operator to specify the visual prompt before executing the task. The operator clicks on the live video stream from the table-mounted camera to generate the visual prompts, which are recorded alongside the robot's observations and actions. The dataset covers the three task types:
\begin{itemize}
    \item \textbf{Displacement ($\mathbf{200}$ episodes):} The workspace contains $1$ to $3$ objects placed at random positions with random orientations. The operator selects a target object (first click) and a desired goal location (second click), then manipulates the robot to push the object to the target.
    \item \textbf{Grouping ($\mathbf{200}$ episodes):} The workspace contains $2$ to $4$ objects placed at random positions with random orientations. The operator clicks two distinct objects, designating the first as the source object to be pushed and the second as the target object. The operator then pushes the first object into proximity with the second.
    \item \textbf{Singulation ($\mathbf{150}$ episodes):} A cluster of $3$ objects is placed in close proximity (touching or near-touching). The operator clicks the target object to be isolated (first click) and a second point indicating the direction of the push. The operator then pushes the object away from the clutter along the specified vector.
\end{itemize}

\section{Experiments and Results}

We design our experiments to evaluate the performance of the proposed Unified Pushing Policy and the prompting mechanism. Specifically, we aim to answer the following questions:
\begin{enumerate}
    \item \textbf{Multi-task:} Can multimodal pushing skills be consolidated into a unified policy with comparable or better performance?
    \item \textbf{Prompt Efficacy:} Does the prompting mechanism offer advantages over dense goal images?
    \item \textbf{Generalization:} Can the policy generalize to novel objects not seen during training?
    \item \textbf{Planner Integration}: Can the policy be used as a low-level primitive in a planning framework?
\end{enumerate}
To answer the first two questions, we conduct experiments comparing our proposed method with baseline models (Sections~\ref{subsec:exp1} and~\ref{section:exp_goal_image}). The third question is addressed by evaluating generalization to unseen objects (Section~\ref{section:exp_unseen}). Finally, we integrate the learned grouping skill into a VLM-driven task planning framework to solve the table-cleaning problem as in~\cite{push_Arxiv2025_Yonemaru} (Section~\ref{section: exp_vlm}). We report the \textbf{Success Rate} for each experiment. A trial is considered successful if a human evaluator judges that the final state satisfies the implicit task-specific constraints described in section \ref{subsec:problem}.

\subsection{Comparison with single-task policies and goal-image policy}
\label{subsec:exp1}
We compare the unified policy against its single-task counterparts as well as a flow matching policy with goal-image conditioning:

\paragraph{Single-Task Policies} We train three separate models with the same architecture as the unified policy, but each trained exclusively on the data for a single task (Displacement-only, Grouping-only, and Singulation-only). 
\paragraph{Goal-Image Policy}  We train a multi-task Flow Matching policy \cite{push_rouxel2024flow}, where the condition is a goal image instead of the prompting mechanism. To obtain the goal images used for evaluation, a human operator performs the task first, and the final state image is captured and used as the goal input for the policy. The goal image is then processed by the same Resnet34 backbone. For classifier-free guidance, we set all pixel values to $0$ to obtain the null goal.

We evaluate the models on $20$ trials per task.  The scenes are initialized with objects in random positions and random orientations: $1$ for displacement task, $2$ objects for grouping task, and a clutter of 3 objects for singulation task. 

\begin{table}[]
\caption{Success Rate (\%) Comparison with baselines ($20$ trials)}
\label{tab:exp1_results}
\centering
\begin{tabular}{lccc}
\toprule
\textbf{Method} & \textbf{Displacement} & \textbf{Grouping} & \textbf{Singulation} \\
\midrule
Goal-Image Policy & $60$ & $60$ & $30$ \\
Single-Task Policy & $75$ & $60$ & $\mathbf{70}$ \\
\textbf{Unified Policy (Ours)} & $\mathbf{85}$ & $\mathbf{70}$ & $65$ \\
\bottomrule
\end{tabular}
\vspace{-8pt}
\end{table}

The results are summarized in Table \ref{tab:exp1_results}. Compared with single-task policies, the unified policy achieves better performance in displacement and grouping $(+10\%)$, while achieving comparable performance in singualtion ($65\%$ versus $70\%$). We hypothesize that learning multiple tasks allows for positive transfer. For instance, the mechanics of pushing an object for displacement are fundamentally similar to pushing an object for grouping, allowing the unified model to learn a more robust policy from a larger, diverse dataset. In addition, our method outperforms the goal-image policy across all tasks. The biggest performance gap was observed in the singulation task, where our method achieves a $65\%$ success rate compared to  $30\%$ for the goal-image baseline. In displacement and grouping, the unified policy maintains a clear advantage ($85\%$ vs. $60\%$ and $70\%$ vs. $60\%$, respectively).

\subsection{Comparison with goal-image policy in cluttered scene}
\label{section:exp_goal_image}
\begin{table}[t]
\caption{Success Rate (\%) in Cluttered Environments ($10$ trials)}
\label{tab:exp2_clutter}
\centering
\begin{tabular}{l c c}
\toprule
\textbf{Task Setting} & \textbf{Unified Policy (Ours)} & \textbf{Goal-Image} \\
\midrule
Displacement ($3$ objs) & $\mathbf{70}$ & $40$ \\
Displacement ($5$ objs) & $\mathbf{50}$ & $30$ \\
Grouping ($3$ objs) & $\mathbf{70}$ & $40$\\
Grouping ($5$ objs) & $\mathbf{70}$ & $40$ \\
Singulation ($2$ clusters) & $\mathbf{50}$ & $40$ \\
\bottomrule
\end{tabular}
\vspace{-8pt}
\end{table}

To evaluate the robustness and effectiveness of the prompting mechanism, we compare the proposed method with the goal-image conditioned policy in cluttered environments. The prompting mechanism is evaluated under increasing scene complexity by introducing distractor objects. Specifically, for the displacement and grouping tasks, we consider two scene settings containing either $3$ or $5$ randomly placed objects in the workspace. For the singulation task, the scene consists of two separate clusters, each composed of $3$ objects.

As shown in Table \ref{tab:exp2_clutter}, our method consistently achieves better results than goal-image policy. In $3$-object scenario, which still falls within training data distribution, the unified policy maintains a high success rate ($70\% $ for displacement and grouping), while the goal-image policy sees a drop in performance to $40\%$. When the scene complexity is increased to $5$ objects (out-of-distribution), we observe a performance decline with the unified policy, dropping to $50\%$ in displacement and singulation tasks. 

We hypothesize that the performance drop of the goal-image policy is due to \textit{over-attention} to the goal-image. To generate a goal image for evaluation. A human demonstrator must first perform the task. In a cluttered environment, it is easy for the demonstrator to accidentally disturb neighbouring objects. These accidental shifts can distract the policy, making it ignore the main task. In contrast, our prompting mechanism provides a more abstract goal representation, which can help mitigate the over-attention problem.

\subsection{Generalization to unseen objects}
\label{section:exp_unseen}
We evaluate our method on a set of novel objects with a similar shape (rectangular boxes of size $115 \times 34 \times 34$ mm), which are slightly larger than the training objects. Bottom right of Fig.\ref{fig:experimental_setting} shows a comparison between two objects. We performed $10$ trials per task, each trial follows the same setting as the experiment in section \ref{subsec:exp1}. 

As a result, the model achieves success rates of $\mathbf{70\%}$, $\mathbf{70\%}$, and $\mathbf{80\%}$ for displacement, grouping, and singulation, respectively. Notably, the success rate for singulation increased compared to the training objects ($65$\% $\to$ $80$\%). Analysis of failure cases reveals that with the original objects, the most common failure case when performing singulation is when the gripper occasionally clips a neighboring object in the cluster, causing an unintended double-push. The slightly larger width dimension of the novel objects mitigates this issue by providing a larger tolerance for the gripper's contact point. This experiment shows that our method can learn generalizable pushing behaviors rather than overfitting to specific object instances.

\subsection{Integration with high-level planning}
\label{section: exp_vlm}
\begin{figure}[t]
    \centering
    \includegraphics[width=1.0\linewidth]{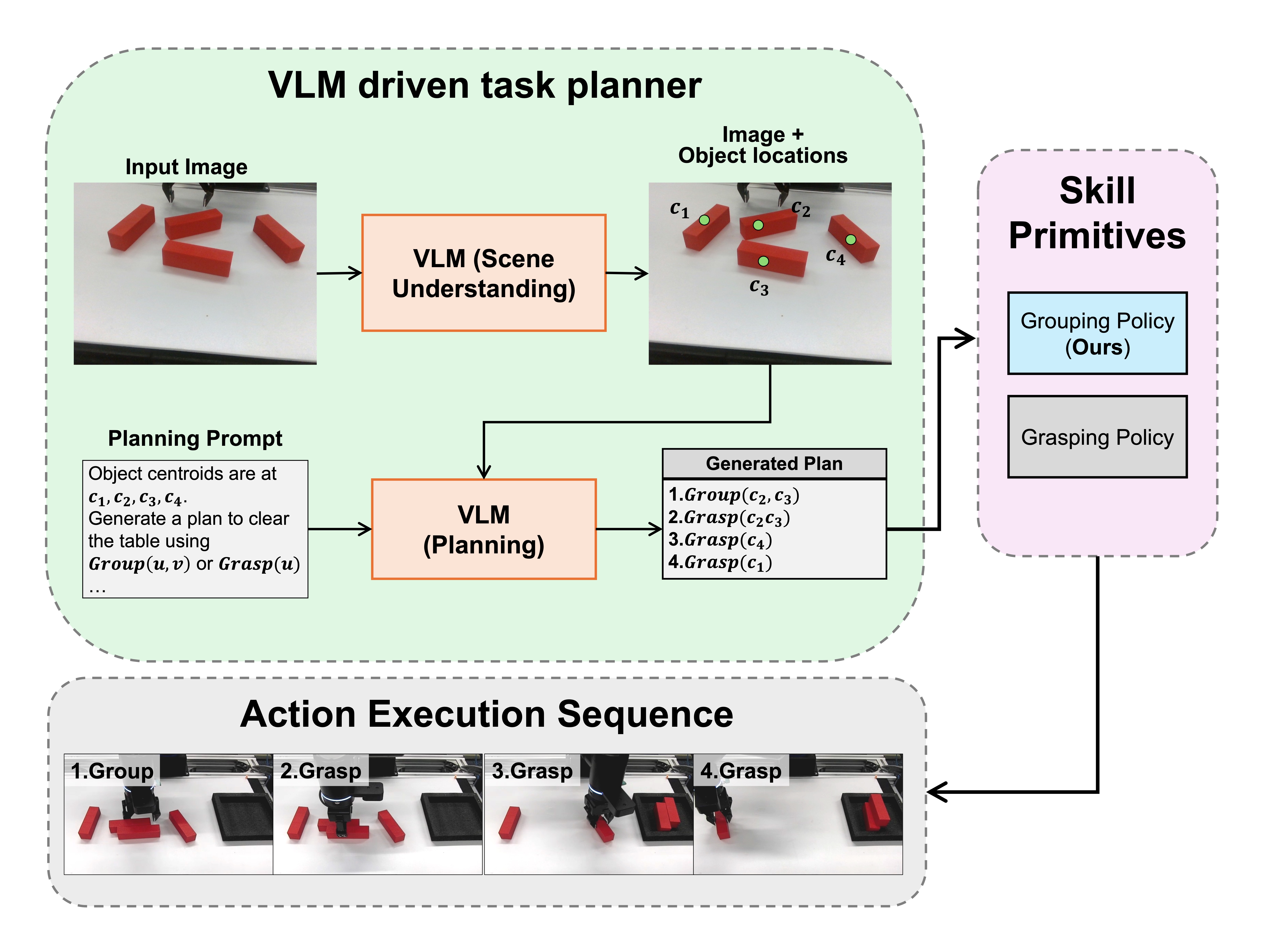}
    \caption{Overview of the VLM planning framework}
    \label{fig:mog}
    \vspace{-10pt}
\end{figure}

\begin{figure*}[t]
    \centering
    \includegraphics[width=1.0\linewidth]{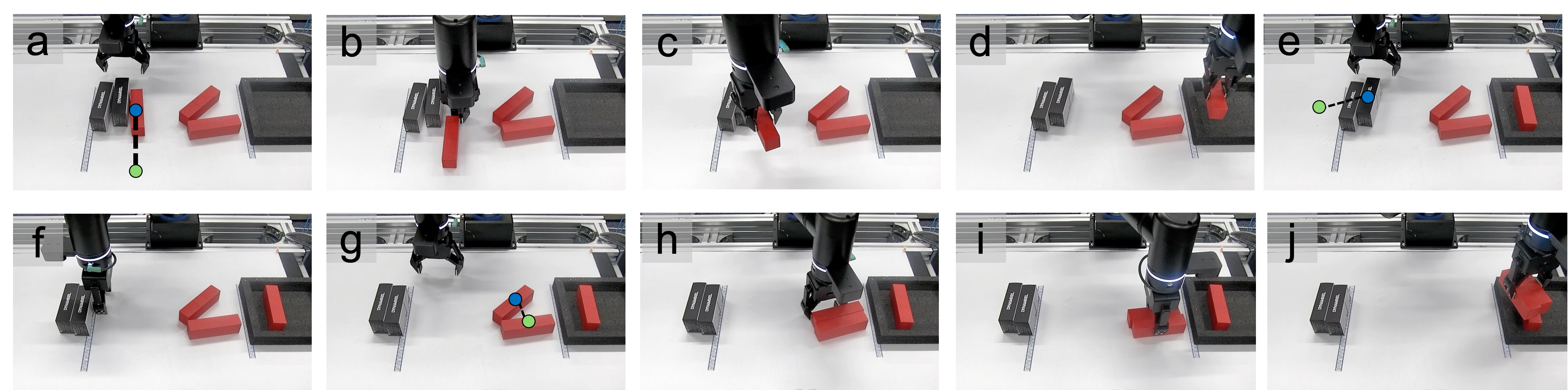}
    \caption{Snapshots of a table-cleaning task with manual prompting. (a)-(d) \textbf{Singulation} to isolate a red block and perform pick-and-place. (e)-(f) \textbf{Displacement} to rearrange $2$ black boxes into target area. (g)-(j) \textbf{Grouping} to group $2$ red blocks and simultaneously pick-and-place them to target regions. Note that the annotated dots in (a), (e), and (g) are the visual prompts: the blue dot represents $\mathbf{u}_1$ while the other represents $\mathbf{u}_2$.}
    \label{fig:real_cleaning}
    \vspace{-10pt}
\end{figure*}
To demonstrate the applicability of our model as a low-level primitive, we develop a naive VLM-driven task planning framework based on Qwen3-VL~\cite{vlm_bai2025qwen3vltechnicalreport}. The objective is to clear a table containing $4$ objects by delivering them to a container using \texttt{Group} and \texttt{Grasp} functions, following the same task setting as in~\cite{push_Arxiv2025_Yonemaru}. Specifically, the \texttt{Group} function is achieved by the unified pushing policy, while \texttt{Grasp} is achieved by a modified Fast Graspability \cite{domae2014fast}, which favors the grasping candidates that can pick multiple objects at once.
Fig. \ref{fig:mog} shows an overview of the pipeline and an exemplary snapshot of the robot executing a generated action sequence. 
Unlike~\cite{push_Arxiv2025_Yonemaru}, where the policy is specifically tailored to the table-cleaning task, our goal is to learn reusable pushing policies that can be flexibly integrated into high-level planners.
The planning process happens in two stages:

\begin{enumerate}
    \item \textbf{Scene Understanding:} The VLM is provided with an RGB image from the table-mounted camera and prompted to output the pixel coordinates of all object centroids.
    \item \textbf{Action Plan Generation:}  We prompt the VLM to generate a sequence of operations to clear the table into a container. The VLM can choose between two primitives: \texttt{Group(target\_1, target\_2)} or \texttt{Grasp(target)}. The planner is required to flexibly employ \texttt{Group} to push near objects closer to facilitate the multi-object grasping.
\end{enumerate}
The robot executes the generated action plan open-loop (without replanning after every step) until all actions are exhausted. 

As a result, we observe a task completion rate of $\mathbf{50\%}$ and an average of \textbf{1.52} objects per grasp action over $10$ trials. The system successfully utilizes the pushing policy to group objects, therefore enabling multi-object grasping to reduce the total number of pick and place operations to clear the scene. The failure cases mainly occur during the grouping stage, where the policy fails to properly align the objects, making the subsequent multi-object grasp only able to secure one object. During data collection, human demonstrators have the tendencies to click near the center of the object to generate the prompt. In contrast, we observe that the coordinate output by the VLM can be any arbitrary point on the object's surface (e.g., near edges or corners). This can potentially lead to suboptimal pushing behavior. Fig.~\ref{fig:real_cleaning} shows that our framework can solve a more difficult table-cleaning task when accurate prompts are provided, highlighting the potential as low-level primitives.



\section{Conclusion and Future Work}
In this work, we proposed a flow matching-based imitation learning policy capable of performing multiple non-prehensible manipulation tasks (object displacement, grouping, and singulation). Real-world experiments show that our method achieves overall better performance than goal-image baseline and single-task policies. Furthermore, we also demonstrated its utility as a low-level primitive within a VLM planning framework, enabling multi-object grasping. 

Despite these results, several areas of improvement remain. Firstly, failure case analysis reveals a frequent issue where the gripper loses contact with objects during pushing due to friction or center-of-mass variations. Since collecting exhaustive data to account for every scenario is impractical, we plan to explore world modeling to capture these dynamics, thereby improving data efficiency. In addition, although we showcased initial compatibility of an imitation learning policy with a naive high-level planner for object grouping, integration with other tasks is currently constrained by the spatial reasoning limitations of off-the-shelf VLMs. We plan to further explore more robust planning frameworks to fully exploit the multimodal capabilities of our method. Finally, the current prompting mechanism relies on a fixed camera viewpoint, so future research will investigate view-invariant prompt representations to enhance deployment flexibility.




\bibliographystyle{IEEEtran}
\bibliography{biblio}
\end{document}